\documentclass[letterpaper]{article} 
\usepackage{aaai24}  
\usepackage{times}  
\usepackage{helvet}  
\usepackage{courier}  
\usepackage[hyphens]{url}  
\usepackage{graphicx} 
\usepackage{natbib}  
\usepackage{caption} 
\usepackage{algorithm}
\usepackage{algorithmic}
\usepackage{amsmath}
\usepackage{amsfonts}

\newtheorem{theorem}{Theorem}

\newtheorem{lemma}[theorem]{Lemma}

\newtheorem{definition}{Definition}

\newtheorem{proposition}{Proposition}
\usepackage{subcaption}
\usepackage{multirow}
\usepackage{newfloat}
\usepackage{listings}
\usepackage{array}
\DeclareCaptionStyle{ruled}{labelfont=normalfont,labelsep=colon,strut=off} 
\floatstyle{ruled}
\newfloat{listing}{tb}{lst}{}
\floatname{listing}{Listing}
\title{AdvST: Revisiting Data Augmentations for Single Domain Generalization}

\author {
    Guangtao Zheng\textsuperscript{\rm 1},
    Mengdi Huai\textsuperscript{\rm 2},
    Aidong Zhang\textsuperscript{\rm 1}
}
\affiliations {
    \textsuperscript{\rm 1}Department of Computer Science, University of Virginia\\
    \textsuperscript{\rm 2}Department of Computer Science, Iowa State University\\
    gz5hp@virginia, mdhuai@iastate.edu, aidong@virginia.edu
}

\usepackage{bibentry}

\begin{document}

\maketitle

\begin{abstract}
Single domain generalization (SDG) aims to train a robust model against unknown target domain shifts using data from a single source domain. Data augmentation has been proven an effective approach to SDG. However, the utility of standard augmentations, such as translate, or invert, has not been fully exploited in SDG; practically, these augmentations are used as a part of a data preprocessing procedure. Although it is intuitive to use many such augmentations to boost the robustness of a model to out-of-distribution domain shifts, we lack a principled approach to harvest the benefit brought from multiple these augmentations. Here,  we conceptualize standard data augmentations with learnable parameters as semantics transformations that can manipulate certain semantics of a sample, such as the geometry or color of an image. Then, we propose Adversarial learning with Semantics Transformations (AdvST) that augments the source domain data with semantics transformations and  learns a robust model with the augmented data. We theoretically show that AdvST essentially optimizes a distributionally robust optimization objective defined on a set of semantics distributions induced by the parameters of semantics transformations. We demonstrate that AdvST can produce samples that expand the coverage on target domain data. Compared with the state-of-the-art methods, AdvST, despite being a simple method, is surprisingly competitive and achieves the best average SDG performance on the Digits, PACS, and DomainNet datasets. Our code is available at https://github.com/gtzheng/AdvST.

\end{abstract}

\section{Introduction}
\label{sec:intro}
Domain generalization \cite{balaji2018metareg,li2017deeper,li2018learning,li2019episodic} aims to learn a model that can generalize well on target (test) domains with unknown distribution shifts using multiple source (training) domains. However, having diverse domains for training is a strong assumption due to various practical considerations, such as data collection budgets or privacy issues. A realistic alternative is \textit{single domain generalization} (SDG) \cite{zhao2020maximum,li2018learning}, which only requires data from a single source domain for model training.  SDG is challenging for deep image classifiers. Although they have achieved impressive performance on benchmarks, they strongly hinge on the implicit assumption that training and test data follow the same distribution. Their performance can drop significantly when there are shifts between training and test data distributions caused by, for example, changes in object appearance or data collection methods. 

Data augmentation is an effective approach to SDG. It augments the source domain data to expand the coverage on the unseen target domain during model training. Methods of data augmentation include using adversarial learning \cite{volpi2018generalizing,zhao2020maximum,qiao2020learning} or using generative models \cite{qiao2020learning,wang2021learning,li2021progressive} to generate diverse data samples. The utility of standard data augmentations, such as scale, or CutOut \cite{devries2017improved}, has not been fully exploited in SDG. In practice, these augmentation methods have been widely used in model training for \textit{in-distribution} generalization. However, their applications in SDG are limited. In most cases, they serve as a part of the data preprocessing procedure in other SDG methods \cite{volpi2018generalizing,zhao2020maximum,qiao2020learning}. Although it is intuitive that applying multiple standard data augmentations to the source domain data can generate diverse samples and hence improve a model's SDG performance, we lack a principled approach to fully realize the benefit brought from multiple standard data augmentations.

Therefore, in this paper, we revisit standard data augmentations for SDG and develop methods that make them a strong competitor in SDG.  We consider the composition of several standard data augmentation as a semantics transformation which can manipulate certain kinds of semantics of a sample, such as the brightness and hue of an image. Normally, standard data augmentations have pre-specified and fixed parameters. Here, we make these parameters learnable in a semantics transformation so that we can tune these parameters to produce  semantically significant variations and bring new styles that are different from the source domain data. With semantics transformations, we can transform data in the source domain to a fictitious one which has large domain shifts from the source and possibly covers data in target domains, yielding favorable SDG performance. 

To learn semantics transformations for SDG, we propose AdvST, an adversarial learning framework that trains a robust model and generates challenging data samples iteratively with mini-max optimization. In the maximization phase, we learn the parameters of semantics transformations so that the samples transformed by semantics transformations maximize the prediction loss of the model. To avoid learning a trivial solution where the information in the source domain samples is completely lost after semantics transformations, we additionally regularize the distance between the source domain samples and the transformed ones in the deep feature space of the model to keep the core features of the source domain data. In the minimization phase, we train the model with the new samples generated by semantics transformations.

We theoretically show that the learning objective of AdvST connects to that of distributionally robust optimization (DRO) \cite{blanchet2019quantifying,gao2022distributionally}. DRO trains a robust model using the worst-case distribution that leads to the worst model performance on an uncertainty set---a set of neighboring distributions with a predefined value of distributional shifts from the training data. Increasing the coverage of the uncertainty set on the target domain data can improve the model's SDG performance. AdvST can be considered as a special form of DRO whose uncertainty set consists of semantics-induced data distributions which are generated by applying semantics transformations to samples from the source distribution. We demonstrate that AdvST can produce samples in the uncertainty set that expand the coverage on the target domain data.

Our method, despite being a simple method utilizing standard data augmentations, is surprisingly competitive in SDG. AdvST consistently outperforms existing state-of-the-art methods in terms of the average SDG performance on three benchmark datasets.

\section{Related Work}
\label{sec:related}
\noindent\textbf{Domain adaptation and generalization.} Domain adaptation methods \cite{french2017self,ganin2015unsupervised,shu2018dirt,li2018adaptive} have been proposed to solve the problem of generalizing to a target domain where the label information is unknown at training time. These methods mainly aim to align the distributions of source and target domains. However, their setups differ from ours since they require access to samples from the target distribution during training. In contrast, domain generalization methods \cite{balaji2018metareg,li2019episodic,pandey2021domain,shankar2018generalizing} do not require samples from the target domain during training. However, they use training samples from multiple domains instead of  one. 

\noindent\textbf{Single domain generalization.} SDG requires no access to target distributions and only one single source domain for training. The general idea is to augment the source domain data, and there are three types of methods. Methods of the first type \cite{devries2017improved,hendrycks2019augmix,zoph2020learning,Cubuk_2020_CVPR_Workshops,lian2021geometry} use traditional data augmentation to improve \textit{in-domain} generalization performance but often fail to generate samples with large domain shifts for out-of-domain generalization. The second type of methods use adversarial data augmentation to augment the source domain data. However, they generate samples either in the pixel space \cite{volpi2018generalizing,zhao2020maximum} or via perturbing latent feature statistics \cite{zhong2022adversarial,zhang2023adversarial}, which struggle to produce samples with large domain shifts. Our method exploits the domain knowledge in standard data augmentations and uses them as semantics transformations with learnable parameters, generating samples with large domain shifts from the source domain. Adversarial AutoAugment \cite{zhang2019adversarial} adversarially learns augmentation policies to improve \textit{in-domain} generalization performance. In contrast, our method directly generates worst-case samples to improve \textit{out-of-domain} generalization performance. The third type \cite{qiao2020learning,wang2021learning,li2021progressive} uses generative models to produce diverse training samples. However, since generative models are learned from the source domain, the styles of the generated samples are still related to those in the source domain. In contrast, our method uses semantics transformations to manipulate the semantics that is \textit{independent} of the source domain, allowing us to inject \textit{external} styles to the generated samples.

\noindent\textbf{Semantics transformations.}
Semantics transformations \cite{mohapatra2020towards} can manipulate certain kinds of the semantics of an image, such as changing hue and saturation \cite{Hosseini2018SemanticAE} or color and texture \cite{Bhattad2020Unrestricted}. Semantics transformations are used to produce ``unrestricted" perturbations \cite{Bhattad2020Unrestricted} in adversarial samples, which are traditionally generated by finding imperceptible perturbations under a norm ball constraint \cite{Bhattad2020Unrestricted}. However, these methods cannot be directly adopted in our problem setting since they focus on performing adversarial attacks, while our goal of using semantics transformations is to improve a model's SDG performance. Semantics transformations have also been used to improve few-shot generalization \cite{zheng2023learning} via meta-learning \cite{finn2017model,zheng2022knowledge}.
A parallel work \cite{gokhale2022semantically} uses a pre-defined set of linguistic transformations, such as negation and paraphrasing, to augment text data for improved vision-language inference performance. However, these transformations do not have learnable parameters and cannot be fine-tuned into different ones.

\section{AdvST: Adversarial Learning With Semantics Transformations}\label{sec:optimization}
\subsection{Semantics Transformation} 
We define a semantics transformation as a composition of several standard data augmentation functions that manipulate certain kinds of semantics of a sample. For example,  we can perturb both the hue and brightness of an image $x$ with $\tau(x;\omega)=o_{h}(o_{b}(x;\omega_{b});\omega_{h})$, where $o_{h}$ is the function that changes the hue of $x$, $o_{b}$ changes the brightness of $x$, and  $\omega=\omega_{b}\cup \omega_{h}$ denotes the set of parameters for $\tau$. We construct a set of $M$ semantics transformations $\mathcal{T}=\{\tau_i(\cdot;\omega_i),i=1,\ldots,M\}$ by randomly composing $L (1\leq L\leq L_{\max})$ unique standard data augmentation functions (see Appendix). 

Intuitively, a semantics transformation with a large $L$ can produce more diverse samples. However, depending on the target domain data, the semantics transformations that produce more diverse samples are not necessarily better than those producing less diverse ones.  Since we have no knowledge about target domains in SDG, we first uniformly choose the length for semantics transformations and then uniformly choose a semantics transformation with the selected length. Thus, we derive the distribution over $M$ semantics transformations as  $G(\tau^L)=\frac{1}{M_LL_{\max}}$,
 where $\tau^L$ denotes a semantics transformation with $L$ standard augmentations, $L_{\max}$ is the maximum number of standard augmentations in $\tau^L$, and $M_L$ is the total number of $\tau^L$ and satisfies $M=\sum_{L=1}^{L_{\max}}M_L$.
\subsection{Learning Objective of AdvST}
SDG aims to train a model that is robust to unseen domain shifts with the training samples from a single source domain. The robustness of the trained model to unseen domain shifts depends on how much the training data covers  target domains. Therefore, with semantics transformations, we aim to generate new data samples that have large domain shifts from the source domain, increasing the chance of covering data samples from unseen target domains. 

A key property of the samples from target domains is that they often yield a high average prediction loss because of their large domain shifts from the source. This motivates AdvST, an adversarial learning framework that learns semantics transformations to generate challenging samples with significant semantics variations for model training. 

Given a model $f_{\theta}$ with parameters $\theta$, a set of source domain samples $\mathcal{D}_S=\{(x_n,y_n)\}_{n=1}^{N}$ with $N$ pairs of training sample $x_n$ and its label $y_n$, and a distribution $G$ over a set of $M$ semantics transformations $\{\tau_i(\cdot;\omega_i)\}_{i=1}^M$, we express the learning objective of AdvST as:
\begin{align}\label{eq:objective}
\theta^*=\min_{\theta\in\Theta}\max_{\psi\in\Psi}\underset{\tau\sim G}{\mathbb{E}}\underset{\xi\sim \mathcal{D}_S}{\mathbb{E}}\big[ \ell(\theta;\xi')-\lambda d_\theta(\xi',\xi)\big],
\end{align}
where $\xi=(x,y)$ denotes a tuple of a sample $x$ and its label $y$, $\xi'=(\tau(x;\omega),y)$ is the tuple of the same label $y$ and a new sample obtained by applying the semantics transformation $\tau$ to $x$,  $\ell(\theta;\xi)$ is the prediction loss for $\xi=(x,y)$, $\Theta$ denotes the set of all possible values of $\theta$, $\psi=\cup_{i=1}^M\omega_i$ denotes the union of the parameters of $M$ semantics transformations, $\Psi$ is the set of all possible values of $\psi$, $\lambda$ is a nonnegative regularization parameter, and $d_{\theta}$ is the squared Euclidean distance function between  $\xi$ and $\xi'$ in the deep feature space of the model $f_{\theta}$,  i.e., $d_{\theta}(\xi,\xi')=\|v-v'\|_2^2$ with the embeddings $v$ and $v'$ of $x$ and $x'$, respectively.

The objective in \eqref{eq:objective} aims to train a robust model with the challenging samples generated from the samples in the source domain while maintaining the core features of the original data.  The novel part of \eqref{eq:objective} is that instead  of generating images in the pixel space, we adversarially learn the parameters of semantics transformations, exploiting the domain knowledge in standard data augmentations to generate diverse images.

\begin{algorithm}[t]
		\caption{Adversarial learning with semantics transformations (AdvST)}
		$\mathbf{Input}$: Source dataset $\mathcal{D}_S$, extended training set $\mathcal{D}$ with $K$ domains, distribution over $M$ semantics transformations $G$, initial model weights $\theta_{0}$,  number of training epochs $E$, batch size $B$, number of batches per epoch $N_B$, and number of updates in the maximization procedure $T_{\max}$\\
		$\mathbf{Output}$: learned weights $\theta$
		\begin{algorithmic}[1]
		\STATE $\theta \leftarrow \theta_{0}$, $\mathcal{D}\text{.add}(\mathcal{D}_S)$
        \FOR{$e=1,\ldots,E$}
        \STATE{//Minimization procedure}
         \FOR{$b=1,\cdots,N_B$}
          \STATE Get a batch of $B$ samples $\mathcal{B}$ from $\mathcal{D}$
          \STATE Update $\theta$ with Eq. \eqref{eq:minimization}
         \ENDFOR
         \STATE{//Maximization procedure}
         \STATE{Initialize an empty $\mathcal{D}_e$}
         \FOR{$(x_n,y_n)\in\mathcal{D}_S$}
         \STATE{Sample $\tau$ from $G$ and initialize its parameters $\omega_{n}^0$}
          \FOR{$t=1,\cdots,T_{max}$}
          \STATE{Generate a sample $x_n^{t}=\tau(x_n;\omega_n^{t-1})$}
           \STATE Update $\omega_n^{t}$ with Eq. \eqref{eq:maximization}
          \ENDFOR
          \STATE Append $(\tau(x_n,\omega_n^{T_{\max}}),y_n)$ to $\mathcal{D}_{e}$
         \ENDFOR
         \STATE{$\mathcal{D}\text{.add}(\mathcal{D}_{e})$}
        \ENDFOR
        \RETURN $\theta$ 
		\end{algorithmic}\label{alg:1}
\end{algorithm}

\subsection{Learning Algorithm}
We adopt an iterative optimization algorithm \cite{volpi2018generalizing,zhao2020maximum} to solve \eqref{eq:objective}. Specifically, the algorithm consists of a minimization and a maximization optimization procedures.

\noindent\textbf{Maximization procedure.} We generate worst-case samples via optimized semantics transformations.
Specifically, we first sample a semantics transformation $\tau$ from $G$. Then, we sample an example $x_n$ from $\mathcal{D}_S$.  We solve the inner maximization problem in (\ref{eq:objective}) by applying $T_{\max}$ steps of stochastic gradient ascent to the parameters of the sampled semantics transformation $\tau$. To facilitate generating diverse samples, we add a maximum entropy regularizer \cite{zhao2020maximum} during the optimization. In the $t$th ($1\leq t\leq T_{\max}$) iteration, we have the following steps:
\begin{align}
 x_n^{t}&=\tau(x_n;\omega_n^{t-1})\label{eq:semantic-augmentation}\\
    \omega_n^{t}& =  \omega_n^{t-1} + \beta \nabla_{\omega_n^{t-1}} \Big(\ell(\theta;x_n^{t},y_n)-\nonumber\\ 
&  \lambda d_{\theta}((x_n^{t},y_n),(x_n,y_n))+\epsilon l_{ent}(\theta;x_n^{t},y_n)\Big), \label{eq:maximization}
\end{align}
where $\omega_n^{t}$ denotes the learnable parameters of $\tau$ for the $n$-th data sample at iteration $t$, $l_{ent}(\theta;x_n^{t},y_n)$ is an entropy regularization term (see Appendix) to further promote learning diverse samples, $\epsilon$ is a nonnegative regularization parameter, and $\beta$ denotes the learning rate in this procedure. We repeat the above steps until all samples in $\mathcal{D}_S$ have been processed. The synthetic data points $\{(\tau(x_n;\omega_n^{T_{\max}}),y_n)\}_{n=1}^N$ are treated as a new domain of data. We add these generated samples to the extended training set denoted as $\mathcal{D}$, which is initialized as $\mathcal{D}_S$.

\noindent\textbf{Minimization procedure.} We use samples generated from the maximization step to train a robust model $\theta$ against unseen distribution shifts. To avoid model forgetting, at each iteration, we sample a batch of $B$ samples $\mathcal{B}$ from the extended training set $\mathcal{D}$ to also use previously generated samples. We add a regularizer $\ell_{reg}(\theta;\mathcal{B})$ consisting of contrastive and entropy loss terms (see Appendix) to facilitate learning robust representations. At each iteration, we update the model parameters $\theta$ using mini-batch stochastic gradient descent as follows 
    \begin{equation}\label{eq:minimization}
     \begin{aligned}
    \theta& \leftarrow \theta -  \alpha\nabla_{\theta}\Big(\frac{1}{B}\sum_{(x,y)\in\mathcal{B}}\ell(\theta;x,y) +\ell_{reg}(\theta;\mathcal{B})\Big),
    \end{aligned}
   \end{equation}
 where ``$\leftarrow$" denotes value assignment, and $\alpha$ denotes the learning rate. The complete algorithm is shown in Algorithm \ref{alg:1}. We further analyze the space and time complexities of the algorithm with practical considerations in the following.

\noindent\textbf{Space complexity.} In the iterative optimization,  we keep adding the generated samples to the extended training set $\mathcal{D}$. The size of $\mathcal{D}$ increases with the iteration number, which is not scalable when the initial training set $\mathcal{D}_S$ or the iteration number is large.\textit{ Therefore, we implement $\mathcal{D}$ as a domain pool that only stores the generated samples from the most recent $K$ runs of the maximization procedures.} In practice, depending on the size of $\mathcal{D}_S$, we set $K$ in the range of 2 to 5 to ensure that we have sufficient samples for training without incurring the scalability issue.\\
\noindent\textbf{Time complexity.} The time complexity of each iteration of the optimization is $N_{B}C_{\mu}+T_{\max}C_{G}N_B$, where $C_{\mu}$ denotes the complexity of updating the model, $C_{G}$ denotes the complexity of updating the parameters of semantics transformations, and $N_B$ denotes the number of training batches. Generally, we have $C_{G}\approx C_{\mu}$ because $C_{G}$ and $C_{\mu}$ both include back-propagating the gradients throughout the whole model, and the number of parameters in semantics transformations is negligible compared to the number of model parameters.  Therefore, the time complexity is $O(ET_{\max}N_B)$, where $E$ is the total iterations (epochs). In practice, to reduce the impact of $T_{\max}$, we could perform the maximization on different batches in parallel or do early stopping  when the difference in loss between consecutive maximization steps is lower than a given threshold.

\section{Theoretical Analysis: Connection to DRO}
\label{sec:method}
\subsection{DRO Formulation}
 The learning objective of SDG can be expressed via DRO \cite{gao2022distributionally} since it does not rely on the notion of a known target distribution. Specifically, DRO chooses a set of probability distributions $\mathcal{U}$ called uncertainty set, and then finds a decision $\theta$ from $\Theta$ that provides the best hedge against $\mathcal{U}$ by solving the following mini-max problem:
\begin{align}
\label{eq:DRO}
    & \min_{\theta \in \Theta} \max_{Q\in\mathcal{U}}\mathbb{E}_{\xi\sim Q}[\ell(\theta;\xi)],
\end{align}
where $\ell(\theta;\xi)$ is the prediction loss with the data-label pair $\xi=(x,y)$, $\Theta$ denotes the set of all possible model parameters, and $\mathcal{U}$ contains distributions that are at most $\delta$-distance away from the source distribution $P$. The uncertainty set, $\mathcal{U}=\{Q|D(P,Q)<\delta\}$, depends on a distance metric $D(\cdot,\cdot)$ and a predefined threshold $\delta>0$. The objective in \eqref{eq:DRO} finds an optimized model under the worst-case distribution $Q^*$ found in $\mathcal{U}$ that maximizes the prediction loss.


\subsection{Semantics-Induced Distribution} 
Given a set of $M$ semantics transformations, a semantics-induced distribution $Q_{\psi}(\xi')$ is defined as follows
\begin{align}\label{eq:semantics-induced-distri}
Q_{\psi}(\xi') = \sum_{\tau_i}G(\tau_i) \int_{\xi}p(\xi'|\tau_i,\xi,\omega_i)dP,
\end{align}
where $\xi'=(x',y')$, $\xi=(x,y)$ is a sample from the source distribution $P$, $\psi=\cup_{i=1}^M\omega_i$ denotes the  parameters of $M$ semantics transformations, and $p(\xi'|\tau_i,\xi,\omega_i)$ is the probability of obtaining $\xi'$ from $\xi$ and the $i$th semantics transformation  $\tau_i$ with parameters  $\omega_i$. We require that transformed samples are still assigned with their
original labels. Therefore, we have $p(\xi'|\tau_i,\xi,\omega_i)=0$ if $y'\neq y$. Moreover, if  $\tau_i$ is a deterministic transformation, then $p(\xi'|\tau_i,\xi,\omega_i)=1$ when $\tau_i(x;\omega_i)=x'$ and $y'=y$ and $p(\xi'|\tau_i,\xi,\omega_i)=0$ otherwise. If $\tau_i$ is a stochastic transformation, then $p(\xi'|\tau_i,\xi,\omega_i)$ follows the distribution of $\tau_i(x;\omega_i)$.
A sample $\xi'$ from $Q_\psi$ can be obtained by first sampling $\xi$ from $P$ with $y=y'$ and $\tau_i$ from $G$, and then obtaining $x'= \tau_i(x;\omega_i)$. Given $G$ and $P$, $Q_\psi$ fully depends on $\psi$. We denote the set of all semantics-induced distributions as $\mathcal{Q}_{\Psi}=\{Q_{\psi}|\psi\in\Psi\}$, where $\Psi$ is the set of all possible parameters $\psi$.


\subsection{Uncertainty Set of AdvST}
The uncertainty set of AdvST consists of semantics-induced distributions $Q_{\psi}$ around the source distribution $P$ to simulate unseen target distributions. These distributions should not deviate too much from the source to avoid hedging against noisy distributions that are not learnable. Hence, we need a proper distance metric $D(\cdot,\cdot)$ to control the distribution shifts. Since semantics transformations create new data samples, we use Wasserstein distances (Definition \ref{def:wasserstein-distance}) as the metric $D$ to allow a data distribution $Q_{\psi}$ to have a different support from that of $P$.

\begin{definition}\label{def:wasserstein-distance} (\textit{Wasserstein distances} \cite{chen2021distributionally,rahimian2019distributionally,kuhn2019wasserstein} )
Let $\Xi$ be a measurable space. Given a transportation cost function $c:\Xi\times \Xi \rightarrow [0,\infty)$, which is nonnegative, lower semi-continuous, and satisfies $c(\xi,\xi)=0$, for probability measures $Q$ and $P$ on $\Xi$, the Wasserstein distance between $Q$ and $P$ is
\begin{align}
\label{eq:distance}
    W_{c}(Q,P)=\inf_{J \in \prod(Q,P)} \mathbb{E}_{(\xi,\xi')\sim J}[c(\xi,\xi')],
\end{align}
where $\prod(Q,P)$ denotes all joint distributions with marginal distributions being $P$ and $Q$.
\end{definition}

 We define the transportation cost function $c$ in the deep feature space \cite{zhao2020maximum,volpi2018generalizing} to include distributions whose samples have large style variations since these samples may still be close to the samples from the source distribution in the deep feature space. To exclude noisy distributions whose data samples change their original labels in the source domain after transformations, we design the cost of moving a source distribution sample to such a sample as infinity. Specifically, the cost function of moving $\xi=(x,y)\sim P$ to $\xi'=(x',y')\sim Q_\psi$ given the model $\theta$ is defined as follows
\begin{align}\label{eq:cost}
    c_{\theta}((x,y),(x',y')) :=  ||v-v'||_{2}^{2}+\infty \cdot 1\{y \neq y'\},
\end{align}
where $v$ and $v'$ are the model-dependent embeddings for $x$ and $x'$, respectively. 
Therefore, the uncertainty set that we consider in AdvST is
\begin{align}\label{eq:uncertainty-set}
    \mathcal{U}_\Psi=\{Q|Q\in\mathcal{Q}_\Psi, W_c(Q,P)<\delta\},
\end{align}
where $\delta$ ($\delta>0$) denotes the predefined distance threshold between the source $P$ and the semantics-induced distributions $\mathcal{Q}_\Psi$. 

\subsection{DRO Learning Objective for AdvST} \label{sec:learning-objective}

Directly solving Eq.~(\ref{eq:DRO}) with $\mathcal{U}=\mathcal{U}_{\Psi}$ is intractable since it requires searching over the infinite dimension space of distribution functions. 
 We consider the following Lagrangian relaxation with the penalty parameter $\lambda$:
\begin{align}
\label{eq:obj-AdvST-lagrange}
    & \min_{\theta \in \Theta}\ \max_{Q\in \mathcal{Q}_\Psi} \{\mathbb{E}_{(x,y)\sim Q}[\ell(\theta;x,y)]-\lambda W_{c}(Q,P)\}. 
\end{align}
 However, Eq. \eqref{eq:obj-AdvST-lagrange} is still hard to compute. For the inner maximization term of Eq. \eqref{eq:obj-AdvST-lagrange}, Proposition \ref{theo:main-theorem} provides a tractable form which only requires the source distribution $P$ and the distribution over semantics transformations $G$.

\begin{proposition}\label{theo:main-theorem}
Let $\ell: \Theta \times \mathcal{X} \times \mathcal{Y} \rightarrow [0,\infty)$ denote the loss function which is upper semi-continuous and integrable. The transportation cost function $c:\Xi\times \Xi \rightarrow [0,\infty)$ with $\Xi=\mathcal{X} \times \mathcal{Y}$ is a lower semi-continuous function satisfying $c(\xi,\xi)=0$ for $\xi\in \Xi$.  Let $G$ denote the distribution over $M$ semantics transformations $\{\tau_i|i=1,\ldots,M\}$. Then, for any given $P$ and $\lambda \geq 0$, it holds that
\begin{equation}
\label{eq:proposition}
\begin{aligned}
    & \sup_{Q \in \mathcal{Q}_\Psi}\{ \mathbb{E}_{Q}[\ell(\theta;x,y)]-\lambda W_{c}(Q,P)\}\\
    &=\mathbb{E}_{\tau_i\sim G}\mathbb{E}_{P}\big[\sup_{\xi \in \Xi_i}(\ell(\theta;\xi)-\lambda c_\theta(\xi,(x,y)))\big].
\end{aligned}
\end{equation}
where $\mathcal{Q}_{\Psi}$ is a set of distributions induced by $M$ semantics transformations parameterized by ${\psi}$,  $\Xi_i=\{(x',y)|x'=\tau_i(x;\omega_i),\xi\in \Xi_0, \omega_i\subset\psi\}$, and $\Xi_0\subseteq \Xi$ is the support of $P$.
\end{proposition}
The proof of Proposition \ref{theo:main-theorem} (see Appendix) includes taking the dual reformulation of Eq. \eqref{eq:obj-AdvST-lagrange} and considering a semantics-induced distribution $Q$ as a mixture of $M$ distributions.
We observe that the objective in \eqref{eq:objective} actually minimizes the empirical version of \eqref{eq:proposition} with $P$ and $c_\theta$ being replaced by $\mathcal{D}_S$ and $d_\theta$, respectively.

\section{Experiment}
\subsection{Experimental Settings}
\noindent\textbf{Datasets.} We use the following three benchmark datasets in the experiments and arrange them in increasing order of difficulty. (1) \textbf{Digits} is used for digit classification and contains five datasets: MNIST \cite{lecun1998gradient}, MNIST-M \cite{ganin2015unsupervised}, SVHN \cite{netzer2011reading}, SYN \cite{ganin2015unsupervised}, and USPS \cite{denker1989neural}. Each dataset has the same 10 digits ranging from 0 to 9.  We use MNIST as the source domain and the other four as the test domains.  (2) \textbf{PACS} \cite{li2017deeper} is a collection of four domains, namely, Art, Cartoon, Photo and Sketch. The four domains share seven common object categories and differ in the styles of their images. We use one domain as the source domain and the other three as the unseen target domains. (3) \textbf{DomainNet} \cite{peng2019moment} is a large-scale dataset which has 345 object classes and contains six domains, namely Real, Infograph, Clipart, Painting, Quickdraw, and Sketch. We use Real as the source domain and the remaining five as the test domains. This is the most challenging dataset in our experiments due to the large number of classes and the high variability of domains.


\noindent\textbf{AdvST implementations.} 
We used 12 standard augmentations commonly used in image transformations (see Appendix), such as Rotate and Translate, to construct semantics transformations.  Most augmentation functions have specific learnable parameters controlling the magnitude of the transformations.  We designed a semantics transformation as a composition of at most $L_{\max}=3$ standard augmentations since more augmentations bring marginal gains.  We used the differentiable library \cite{eriba2019kornia} to implement these transformations.  We denote our method as \textbf{AdvST} when $\epsilon=0$ in Eq. \eqref{eq:maximization} and \textbf{AdvST-ME} when $\epsilon>0$.

\noindent\textbf{Training details.}  (1) Experiments on \textbf{Digits}: we adopted the LeNet \cite{lecun1998gradient} as the backbone and used the first $10,000$ images in MNIST to train the model.   All images are resized to $32 \times 32$ and converted to RGB images. For our method, we set $E=50$, $T_{\max}=20$, $\lambda=100$, $\beta=0.2$, $B=32$, and $\alpha=1\times 10^{-4}$ which is dropped by 0.1 after 25 epochs. 
(2) Experiments on \textbf{PACS}: we used a ResNet-18 \cite{he2016deep} pre-trained on ImageNet as the backbone and fine-tuned it on the source domain. All images are resized to $224 \times 224$. We set $E=50$, $T_{\max}=50$, $\lambda=10$, $\beta=5.0$, $B=32$, and $\alpha=0.001$ which decays following a cosine annealing scheduler. 
(3) Experiments on \textbf{DomainNet}:  we used the same backbone as for the PACS datasets. We set $E=200$, $T_{\max}=50$, $\lambda=10$, $\beta=1.0$, $B=128$, and $\alpha=0.001$ which decays following a cosine annealing scheduler.  We did not specifically tune hyperparameters as our method is robust to them as long as the training converges.  Additional training details are in Appendix. 

We ran our experiments on Nvidia Quadro RTX 8000 GPUs.  We ran our experiments 5 times with different random seeds and reported the average accuracy with standard deviation.  

\begin{table}[t]
\centering
\small
\begin{tabular}{ccccc}
\hline
Config. & Semantics & Contrastive & Entropy  & Avg. \\ \hline
1 & &  &  & 59.3$\pm$1.5 \\
2 & $\checkmark$ &   &  & 77.0$\pm$0.4 \\
3& $\checkmark$& $\checkmark$ &  & 77.8$\pm$0.2 \\
4& $\checkmark$& & $\checkmark$ & 78.8$\pm$0.2 \\
5& $\checkmark$& $\checkmark$ &  $\checkmark$& 80.0$\pm$0.4 \\ \hline
\end{tabular}%
\caption{Ablation study on the Digits dataset. We report average classification accuracy over the four target domains.}
\label{tab:ablation-digits}
\end{table}

\begin{figure}[t]
    \centering
    \includegraphics[width=\linewidth]{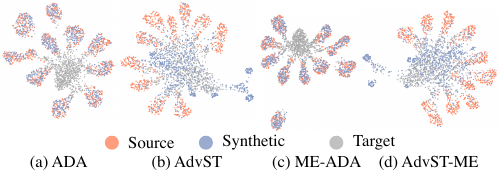}
    \caption{Visualization of how samples from the source domain, target domains, and synthetic domains distribute in the embedding space. We compare AdvST and AdvST-ME with their non-semantics counterparts ADA and ME-ADA.}
    \label{fig:visualization-synthetic}
\end{figure}
\subsection{Ablation Studies}
We conducted ablation studies on AdvST-ME using the Digits dataset. We evaluated how semantics transformations (Semantics), the contrastive regularizer (Contrastive), and the entropy regularizer (Entropy) affect the average SDG performance. We observe from Table~\ref{tab:ablation-digits} that semantics transformations can significantly boost the average classification accuracy by 17.7\% (Configurations 1 and 2).  The contrastive and entropy regularizers can further boost the performance of Configuration 1 by 0.8\% (Configuration 3) and 1.8\% (Configuration 4), respectively.  Our method (Configuration 5) achieves the highest average classification accuracy with all three components.

We further compared the coverage of generated samples on target domain data between our methods, AdvST and AdvST-ME, and their pixel-level counterparts, ADA \cite{volpi2018generalizing} and ME-ADA \cite{zhao2020maximum}, which directly generate images in the pixel space. We visualized how the samples generated by ADA, ME-ADA, and our methods distribute in the embedding space in Figure~\ref{fig:visualization-synthetic}. We color the samples from the source domain MNIST orange and the samples from the four target domains gray. We give details for obtaining the figure in Appendix. From Figure~\ref{fig:visualization-synthetic}(a) and (c), we observe that most of the synthetic samples distribute very close to the source domain data and have little coverage on the target domains. In contrast, the synthetic samples in Figure~\ref{fig:visualization-synthetic}(b) and (d) deviate from the source domain and have broad coverage on the target domains. 

We provide analyses on the sensitivity of $\lambda$  and the effect of different semantics transformations in  Appendix.

\subsection{Comparison on Digits}
\textbf{Baselines.} We included ADA, ME-ADA, and the following methods for comparison: ERM, which trains a model only using the standard cross-entropy loss; CCSA \cite{motiian2017unified}, which aligns samples from different domains to improve generalization; d-SNE \cite{xu2019d}, which minimizes the maximum distance between sample pairs of the same class and maximizes the minimum distance among sample pairs of different categories; JiGen \cite{carlucci2019domain}, which is a multi-task learning method that combines the target recognition task and the Jigsaw classification task; M-ADA \cite{qiao2020learning}, which uses generative models and meta-learning \cite{finn2017model,chen2021hetmaml,chen2022topological} to improve ADA; AutoAug \cite{cubuk2018autoaugment} and RandAug \cite{cubuk2020randaugment}, which augment data based on the searched augmentation policies; RSDA \cite{volpi2019addressing}, which randomly searches image transformations to train a robust model; and PDEN \cite{li2021progressive} and L2D \cite{wang2021learning}, which use generative models for data augmentation.

\begin{table}[t]
\centering
\small
\begin{tabular}{p{1.465cm}@{\hskip 0.1cm}>{\centering}p{0.75cm}@{\hskip 0.5cm}>{\centering}p{1.4cm}@{\hskip 0.05cm}>{\centering}p{0.9cm}>{\centering}p{1cm}p{1cm}}
\hline
Method & SVHN & MNIST-M & SYN & USPS & Avg. \\ \hline
 ERM & 27.8 & 52.7 & 39.7 & 76.9 & 49.3 \\
 CCSA& 25.9 & 49.3 & 37.3 & 83.7 & 49.1\\
d-SNE & 26.2 & 51.0 & 37.8 & 93.2 & 52.1\\
JiGen & 33.8 & 57.8 & 43.8 & 77.2 & 53.1\\
ADA & 35.5 & 60.4 & 45.3 & 77.3 & 54.6\\
 ME-ADA & 42.6 & 63.3 & 50.4 & 81.0 & 59.3\\ 
M-ADA & 42.6 & 67.9 & 49.0& 78.5 & 59.5\\
AutoAug & 45.2 & 60.5 & 64.5 & 80.6 & 62.7 \\
RandAug & 54.8 & 74.0 & 59.6 & 77.3 & 66.4 \\
RSDA & 47.7 & 81.5 & 62.0 & 83.1 & 68.5\\
 L2D & 62.9 &  \textbf{87.3} & 63.7 & 84.0 & 74.5\\ 
  PDEN & 62.2 & 82.2 & 69.4 & 85.3 & 74.8\\ \hline
    AdvST &  \textbf{67.5}$\boldsymbol{\pm}$\textbf{0.7} & 79.8$\pm$0.7 & \textbf{78.1}$\boldsymbol{\pm}$\textbf{0.9} & 94.8$\pm$0.4 &  \textbf{80.1}$\boldsymbol{\pm}$\textbf{0.5} \\
  AdvST-ME & 66.7$\pm$1.0 & 80.0$\pm$0.5 & 77.9$\pm$0.7 & \textbf{95.4}$\boldsymbol{\pm}$\textbf{0.4} & 80.0$\pm$0.4 \\\hline
\end{tabular}%
\caption{Classification accuracy (\%) results on the four target domains SVHN, MNIST-M, SYN, and USPS, with MNIST as the source domain. Best results are in bold font.}
\label{tab:digits}
\end{table}

\begin{table*}[t]
\centering
\small
\begin{tabular}{c|cccccccc|cc}
\hline
Target & MixUp & CutOut & ADA & ME-ADA & AugMix & RandAug & ACVC& L2D & AdvST & AdvST-ME \\ \hline
Art & 52.8 & 59.8 & 58.0 & 60.7 & 63.9 & 67.8 & 67.8 & 67.6 & \textbf{69.2}$\boldsymbol{\pm}$\textbf{1.4} & 67.0$\pm$1.1 \\
Cartoon & 17.0 & 21.6 & 25.3 & 28.5 & 27.7 & 28.9 & 30.3 & 42.6 & \textbf{55.3}$\boldsymbol{\pm}$\textbf{2.0} & 53.2$\pm$1.1 \\
Sketch & 23.2 & 28.8 & 30.1 & 29.6 & 30.9 & 37.0 & 46.4 & 47.1 & \textbf{67.7}$\boldsymbol{\pm}$\textbf{1.5} & 67.2$\pm$2.2 \\
Avg. & 31.0 & 36.7 & 37.8 & 39.6 & 40.8 & 44.6 & 48.2 & 52.5 & \textbf{64.1}$\boldsymbol{\pm}$\textbf{0.4} & 62.5$\pm$0.8 \\ \hline
\end{tabular}%
\caption{Classification accuracy (\%) comparison on the PACS dataset. Best results are in bold font.}
\label{tab:pacs-experiment}
\end{table*}

\begin{table*}[thbp]
\centering
\small
\begin{tabular}{c|cccccccc|cc}
\hline
Target    & MixUp & CutOut & CutMix & ADA  & ME-ADA & RandAug & AugMix & ACVC          & AdvST        & AdvST-ME                                    \\ \hline
Painting  & 38.6  & 38.3   & 38.3   & 38.2 & 39.0   & 41.3    & 40.8   & 41.3          & 42.3$\pm$0.1 & \textbf{42.4}$\boldsymbol{\pm}$\textbf{0.2} \\
Infograph & 13.9  & 13.7   & 13.5   & 13.8 & 14.0   & 13.6    & 13.9   & 12.9          & 14.8$\pm$0.2 & \textbf{14.9}$\boldsymbol{\pm}$\textbf{0.1} \\
Clipart   & 38.0  & 38.4   & 38.7   & 40.2 & 41.0   & 41.1    & 41.7   & \textbf{42.8} & 41.5$\pm$0.4 & 41.7$\pm$0.2                                \\
Sketch    & 26.0  & 26.2   & 26.9   & 24.8 & 25.3   & 30.4    & 29.8   & 30.9          & 30.8$\pm$0.3 & \textbf{31.0}$\boldsymbol{\pm}$\textbf{0.2} \\
Quickdraw & 3.7   & 3.7    & 3.6    & 4.3  & 4.3    & 5.3     & 6.3    & \textbf{6.6}  & 5.9$\pm$0.2  & 6.1$\pm$0.2                                 \\
Avg.      & 24.0  & 24.1   & 24.2   & 24.3 & 24.7   & 26.3    & 26.5   & 26.9          & 27.1$\pm$0.2 & \textbf{27.2}$\boldsymbol{\pm}$\textbf{0.1} \\ \hline
\end{tabular}
\caption[DomainNet result]{Classification accuracy (\%) comparison on the DomainNet dataset. Best results are in bold font.}
\label{tab:domainnet}
\end{table*}

\begin{figure}[tbhp]
  \begin{center}
    \includegraphics[width=0.52\linewidth]{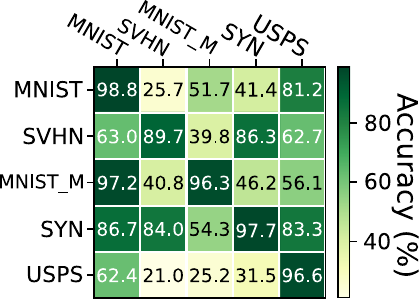}
  \end{center}
 \caption{Accuracy heatmap for models trained individually on the five domains from the Digit dataset using ERM.}
    \label{fig:digits-heatmap}
\end{figure}
\noindent\textbf{Results.} We observe from Table \ref{tab:digits} that our methods, AdvST and AdvST-ME, significantly improve the performance of the pixel-level adversarial data augmentations, ADA and ME-ADA,  across the four target domains and achieve a maximum gain of 25.5\% in 
average classification accuracy. 
Regarding per-domain performance, our methods achieve the best performance on all the target domains except the MNIST-M domain. It is common to observe that a method does not perform the best on all the target domains. For example,  PDEN performs better than L2D on SYN but worse than L2D on MNIST-M. We reason that the knowledge that helps a model generalize in one domain does not necessarily work for the other. To demonstrate this, we trained models on one of the five domains and evaluated their generalization performance on each of the remaining domains. From the accuracy heatmap in Figure~\ref{fig:digits-heatmap}, we see that the learned knowledge for MNIST-M cannot transfer well to SYN and vice versa, which explains the performance tradeoff between MNIST-M and SYN when comparing AdvST with AdvST-ME or AdvST with L2D. Nevertheless, our methods achieve the best average classification accuracy over the four target domains among all the methods.

\subsection{Comparison on PACS }

\noindent\textbf{Baselines.} We compared our methods AdvST and AdvST-ME with ADA, ME-ADA, MixUp \cite{zhang2018mixup}, CutOut \cite{devries2017improved}, CutMix \cite{Yun_2019_ICCV}, RandAug \cite{Cubuk_2020_CVPR_Workshops}, AugMix \cite{hendrycks2019augmix}, and L2D  \cite{wang2021learning}. We also included ACVC \cite{Cugu_2022_CVPR}, which applies attention consistency to learning from augmented samples.

\noindent\textbf{Results.} We used Photo as the source domain and evaluated models on the Art, Cartoon, and Sketch domains. Generalizing raw images to artificial images is the most challenging SDG setting in the PACS dataset since the domain shift between the source and target domains is substantial. Results in Table~\ref{tab:pacs-experiment} show that our methods significantly improve the performance of pixel-level adversarial data augmentations, ADA and ME-ADA, in all three domains. Moreover, our method AdvST performs the best on the three target domains and achieves the best average classification accuracy over the three domains. AdvST-ME performs the second best in this setting, indicating that maximizing output entropy to further encourage generating diverse samples does not help the generalization from a natural domain to an artificial one.

\subsection{Comparison on DomainNet }


\noindent\textbf{Baselines.} We compared our methods AdvST and AdvST-ME with ADA, ME-ADA, MixUp \cite{zhang2018mixup}, CutOut \cite{devries2017improved}, CutMix \cite{Yun_2019_ICCV}, RandAug \cite{Cubuk_2020_CVPR_Workshops}, and AugMix \cite{hendrycks2019augmix}.

\noindent\textbf{Results.} Table \ref{tab:domainnet} shows our results in the most challenging SDG setting, DomainNet, which has 345 classes and significant domain shifts from the source domain, such as Real to Infograph and Real to Quickdraw. Under this challenging setting, our methods outperforms pixel-level adversarial data augmentations, ADA and ME-ADA, and complex data augmentations, such as RandAug and AugMix.

\begin{figure}[thbp]
    \centering
    \includegraphics[width=0.6\linewidth]{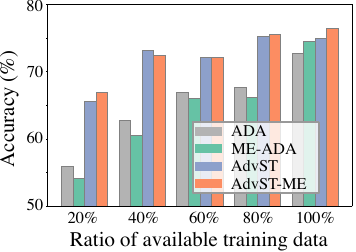}
    \caption{Average classification accuracy under different ratios of available training data. }
    \label{fig:pacs-ratios}
\end{figure}
\subsection{Learning With Limited Source Data}
We further demonstrated the utility of our methods by evaluating the average classification accuracy of our methods on target domains with limited training data. We used the Art dataset from PACS as the source domain and the remaining three datasets in PACS as the target domains. We used partial training data of the Art domain and reported the average classification accuracy over the three target domains in Figure~\ref{fig:pacs-ratios}. We observe that under different ratios of available training data, our methods, AdvST and AdvST-ME, consistently outperform ADA and ME-ADA, respectively. The gains are significant when the ratio is small, demonstrating the effectiveness of our method when there is a lack of available training data.

\section{Conclusion}
We revisited data augmentation for SDG and focused on leveraging the domain knowledge in standard data augmentations. We conceptualized a composition of several standard data augmentations as a semantics transformation with learnable parameters and proposed AdvST, an adversarial learning framework that aims to train a robust model with diverse samples generated by semantics transformations. We theoretically showed that AdvST optimizes a DRO objective with semantics-induced distributions. Although built on standard data augmentations, AdvST is surprisingly competitive. It achieves the best average domain generalization performance on three benchmark datasets and is effective with limited source data. A promising future improvement is to expand the pool of standard data augmentations and selectively choose augmentations given the partial knowledge of target domain data, such as style descriptions.

\section{Acknowledgments}
 This work is supported in part by the US National Science Foundation under grants 2217071, 2213700, 2106913, 2008208, 1955151.
\bibliography{aaai24}
\renewcommand{\thetable}{A\arabic{table}}
\renewcommand{\thefigure}{A\arabic{figure}}
\renewcommand{\theequation}{A\arabic{equation}}
\setcounter{table}{0}
\setcounter{equation}{0}
\setcounter{theorem}{0}
\clearpage
\appendix
\section{Appendix}

\subsection{Proof of Proposition 1}\label{sec:proof-of-theorem1}
We first show that the inner maximization in Eq. \eqref{eq:obj-AdvST-lagrange} satisfies strong duality condition \cite{blanchet2019quantifying} and that the dual problem involves optimization over a one-dimensional dual variable. Lemma \ref{lemma:duality} gives the useful result for \textit{any} distribution $\mathcal{Q}$ satisfying $W_{c}(Q,P) \leq \delta$.  We omit the proof since it is a minor adaptation of Proposition 1 in \cite{blanchet2019quantifying}.
\begin{lemma}\label{lemma:duality}
Let $\ell: \Theta \times \mathcal{X} \times \mathcal{Y} \rightarrow [0,\infty)$ denote the loss function which is upper semi-continuous and integrable. The transportation cost function $c:\Xi\times \Xi \rightarrow [0,\infty)$ with $\Xi=\mathcal{X} \times \mathcal{Y}$ is a lower semi-continuous function satisfying $c(\xi,\xi)=0$ for $\xi\in \Xi$. For any distribution $Q$ and any $\delta \geq 0$, let $s_{\lambda}(\theta;(x,y))=\sup_{\xi \in \Xi}(\ell(\theta;\xi)-\lambda c(\xi,(x,y)))$. Then, for any given $P$ and $\delta > 0$, it holds that
\begin{align}
    & \sup_{Q \in \mathcal{Q}} \mathbb{E}_{Q} [\ell(\theta;x,y)]=\inf_{\lambda \geq 0} \{\lambda \delta+\mathbb{E}_{P}[s_{\lambda}(\theta;(x,y))]\}
\end{align}
and for any $\lambda \geq 0$, we have
\begin{align}
\label{eq:Theorem22}
    & \sup_{Q \in \mathcal{Q}} \{\mathbb{E}_{Q} [\ell(\theta;x,y)]-\lambda W_{c}(Q,P)\} =\mathbb{E}_{P}[s_{\lambda}(\theta;(x,y))].
\end{align}
where $\mathcal{Q}=\{Q: W_{c}(Q,P) \leq \delta\}$.
\end{lemma} 

Note that in our AdvST framework, the distribution $Q$ is semantics-induced and is defined as a mixture of $M$ distributions as shown in Eq.~\eqref{eq:semantics-induced-distri}. To get a tractable learning objective through Lemma \ref{lemma:duality}, let $Q_i=\int_{\xi}p(\xi'|\tau_i,\xi,\omega_i)dP$, and we have the following
\begin{align}
    &\mathbb{E}_{Q_\psi}[\ell(\theta;x,y)]-\lambda W_c(Q_\psi,P)\\
&=\mathbb{E}_{\tau_i\sim G}\big[\mathbb{E}_{Q_i}\big(\ell(\theta;x,y)\big)\big]-\lambda \mathbb{E}_{\tau_i\sim G}[W_c(Q_i,P)]\\
& = \mathbb{E}_{\tau_i\sim G}\big[\mathbb{E}_{Q_i}\big(\ell(\theta;x,y)\big)-\lambda W_c(Q_i,P) \big]\label{eq:max-expansion}\\
&=\mathbb{E}_{\tau_i\sim G}\mathbb{E}_{P}\big[\sup_{\xi \in \Xi_i}(\ell(\theta;\xi)-\lambda c_\theta(\xi,(x,y)))\big] \label{eq:max-result},
\end{align}
where Eq.~\eqref{eq:max-result} is the result of applying  Lemma~\ref{lemma:duality} to the inner term of Eq.~\eqref{eq:max-expansion}, and $\Xi_i=\{(x',y)|x'=\tau_i(x;\omega_i),\xi\in \Xi_0, \omega_i\subset\psi\}$ is the support of $Q_i$ with $\Xi_0$ being the support of $P$. Finding the supreme over $\Xi_i$ that maximizes the inner term in Eq.~\eqref{eq:max-result} is equivalent to finding $\omega_i$.

\subsection{Experimental Details}\label{sec:experimental-details}
\paragraph{Semantics transformations.}
The semantics transformations used in the experiments are constructed from the 12 standard image transformations described in Table~\ref{tbl:semantics}. These standard transformations are designed with domain knowledge about image transformations. Each standard transformation manipulates a particular kind of semantics of an image with a few learnable parameters controlling the transformation magnitude. For example, HSV perturbs an image in the HSV color space with three learnable parameters, and Translate changes an object's position in an image with two learnable parameters. Some standard transformations do not have any learnable parameters because they do not need any parameters, such as Equalize, or they are just non-differentiable functions, such as Posterize. For the latter case, we randomly sample values for parameters from their valid ranges and treat the corresponding function as an identity function during back-propagation. We design a semantics transformation as the concatenation of $L_{\max}$ ($L_{\max}=3$ in the experiments) standard transformations to manipulate multiple kinds of semantics in an image.  
\begin{table*}[htbp]
\vspace{-3mm}
\centering
\caption{Standard data augmentations used in experiments.}\label{tbl:semantics}
\begin{tabular}{clc}
\hline
\multicolumn{1}{c}{Standard transformations} & \multicolumn{1}{c}{Description} & \begin{tabular}[c]{@{}c@{}}Number of \\ Parameters\end{tabular} \\ \hline
  HSV & Perturb in the HSV color space & 3 \\ \cline{1-3} 
  Contrast & Perturb the contrast of an image & 1 \\ \cline{1-3} 
  Invert & \begin{tabular}[c]{@{}l@{}}Invert pixel values at a \\ given threshold\end{tabular} & 1 \\ \cline{1-3} 
  Sharpness & \begin{tabular}[c]{@{}l@{}}Perturb the sharpness \\ of an image\end{tabular} & 1 \\ \hline
 Shear & \begin{tabular}[c]{@{}l@{}}Shear an image in horizontal \\ and vertial directions\end{tabular} & 2 \\ \cline{1-3} 
 Translate & \begin{tabular}[c]{@{}l@{}}Move an image in horizontal \\ and vertial directions\end{tabular} & 2 \\ \cline{1-3} 
 Rotate & Rotate an image & 1 \\ \hline
 Scale & Change the size of an image  & 1 \\ \hline
 Solarize & Reverse the tone of an image  & 1 \\ \hline
 Equalize &  \begin{tabular}[c]{@{}l@{}}Improve global contrast of\\ an image via equalization\end{tabular} & None \\ \hline
 Posterize & \begin{tabular}[c]{@{}l@{}}Reduce the number of bits \\for each color channel\end{tabular} & None \\ \hline
 Cutout & Produce occlusions in an image & None \\ \hline
\end{tabular}%
\end{table*}

\paragraph{Contrastive regularizer.} The regularizer uses a contrastive loss to facilitate learning domain-invariant features from samples in $\mathcal{D}$, which stores generated samples. The loss ensures that samples with the same label are moved close to each other, and those with different labels are moved away from each other. Concretely, we denote the index of a sample in a batch as  $i$, the set of all indexes as $\mathcal{I}_{\mathcal{B}}$, the index set excluding $i$ as $\mathcal{I}_{\mathcal{B}}(i)=\mathcal{I}_{\mathcal{B}}\backslash \{i\}$, and the indexes of samples with label $y_i$ as $\mathcal{P}(i)=\{p\in \mathcal{I}_{\mathcal{B}}(i)|y_i=y_p\}$. Then, our contrastive regularizer is given as follows:
\begin{align}\label{eq:sup-con}
 \ell_{sc}(\theta;\mathcal{B}) = \sum_{i\in \mathcal{I}_{\mathcal{B}}} \frac{-1}{|\mathcal{P}(i)|} \sum_{p\in \mathcal{P}(i)} \log\frac{\exp(u_i^Tu_p)}{\sum_{a\in \mathcal{I}_\mathcal{B}(i)}\exp(u_i^Tu_a)},
\end{align}
where $u_i=\phi(v_i)$ is the projection of the embedding $v_i$ of the input $x_i$ with $v_i=f_{\theta}(x_i)$, and $\phi$ is a projection function. Choices of $\phi$ are given in the experimental details for specific datasets.

\paragraph{Entropy regularizer.} The regularizer uses output entropy to penalize overly confident predictions and to learn good decision boundaries that benefit model generalization. Specifically, it  calculates the average output entropy for a batch of samples $\mathcal{B}$ as follows
\begin{align}\label{eq:entropy}
   \ell_{ent}(\theta;\mathcal{B})=\frac{1}{|\mathcal{B}|}\sum_{i\in \mathcal{I}_\mathcal{B}}\sum_{j=1}^C- p_{ij}\log p_{ij},
\end{align}
where $\mathcal{I}_\mathcal{B}$ is the index set of samples in $\mathcal{B}$, $C$ is the number of classes/outputs, $p_{ij}$ is the $j$th element of $p_i=\text{softmax}(f_{\theta}(x_i))$. By definition, a confident prediction, which  has a very high value on a particular class, has a low output entropy; while a less confident prediction has a larger output entropy. For a single sample, we have $l_{ent}(\theta;x_i,y_i)=\sum_{j=1}^C- p_{ij}\log p_{ij}$.

The regularizer used in the minimization  (Eq. \eqref{eq:minimization}) $\ell_{reg}(\theta;\mathcal{B})$ is defined as the combination of a contrastive and entropy loss terms, i.e., $\ell_{reg}(\theta;\mathcal{B})=\ell_{sc}(\theta;\mathcal{B}) - \eta \ell_{ent}(\theta;\mathcal{B})$, where $\eta$ is a nonnegative regularization parameter.


\paragraph{Experiments on Digits.} The backbone network (i.e., the whole model except the last classification layer) has two $5\times 5$ convolutional layers. The two layers have 64 and 128 channels, respectively. Each convolutional layer is followed by a $2\times2$ max pooling layer. After the two convolutional layers, there are two fully-connected layers with a size of 1024. The classification layer is a linear layer with 1024 inputs and 10 outputs. To calculate the contrastive loss, we design the projection function $\phi$ in Eq.~\eqref{eq:sup-con} as a linear layer (1024 inputs and  128 outputs) followed by a normalization layer. Specifically, $\phi$ first projects the embeddings from the backbone network and then normalizes the projections to have a unit length. We set $\eta=10$ and $\epsilon=10$. To speed up training, we adopt early stopping in the maximization procedure, i.e., if the difference between the previous loss and current loss is smaller than 0.1, then we exit the maximization procedure.
 
\paragraph{Experiments on PACS.} We use a ResNet-18 \cite{he2016deep} as the backbone network. To facilitate knowledge transfer, we pre-train the network on ImageNet and fix the batch normalization statistics of all its batch normalization layers during fine-tuning. The classification layer is a linear layer with 512 inputs and 7 outputs. To calculate the contrastive loss, we design the projection function $\phi$ in Eq.~\eqref{eq:sup-con} as a linear layer with 512 inputs and  128 outputs followed by a normalization layer. We set $\eta=0.1$ and $\epsilon=1$. We adopt the same early stop technique in the maximization procedure as in the Digits experiments.

\paragraph{Experiments on DomainNet.} We use a ResNet-18 \cite{he2016deep} as the backbone network. To speed up training, we use a ResNet-18 network pre-trained on ImageNet as the initialization for the backbone network. The classification layer is a linear layer with 512 inputs and 345 outputs. To calculate the contrastive loss, we design the projection function $\phi$ in Eq.~\eqref{eq:sup-con} as a linear layer with 512 inputs and  128 outputs followed by a normalization layer. We set $\eta=10$ and $\epsilon=10$. We adopt the same early stop technique in the maximization procedure as in the PACS experiments.

\paragraph{Method for obtaining Figure~\ref{fig:visualization-synthetic}.} We first train a model using ERM on the MNIST domain. The backbone (all layers before the last classification layer) of the trained model will be used to get embeddings of all the sampled images.  Then, we sample 1000 images from the source domain MNIST, the four target domains, and the generated images obtained by a DRO-based method, respectively. We get the embeddings of all the samples using the backbone network. To visualize the embeddings, we use UMAP \cite{mcinnes2018umap} with \texttt{n\_neighbors=100} and 
\texttt{min\_dist=0.9} to get two-dimensional representations of the embeddings.

\subsection{Additional Experimental Results}\label{sec:additional-results}
\subsubsection{Sensitivity Analysis on $\lambda$} 
\begin{figure}[h]
     \centering
     \begin{subfigure}[b]{0.4\linewidth}
         \centering
         \includegraphics[width=\linewidth]{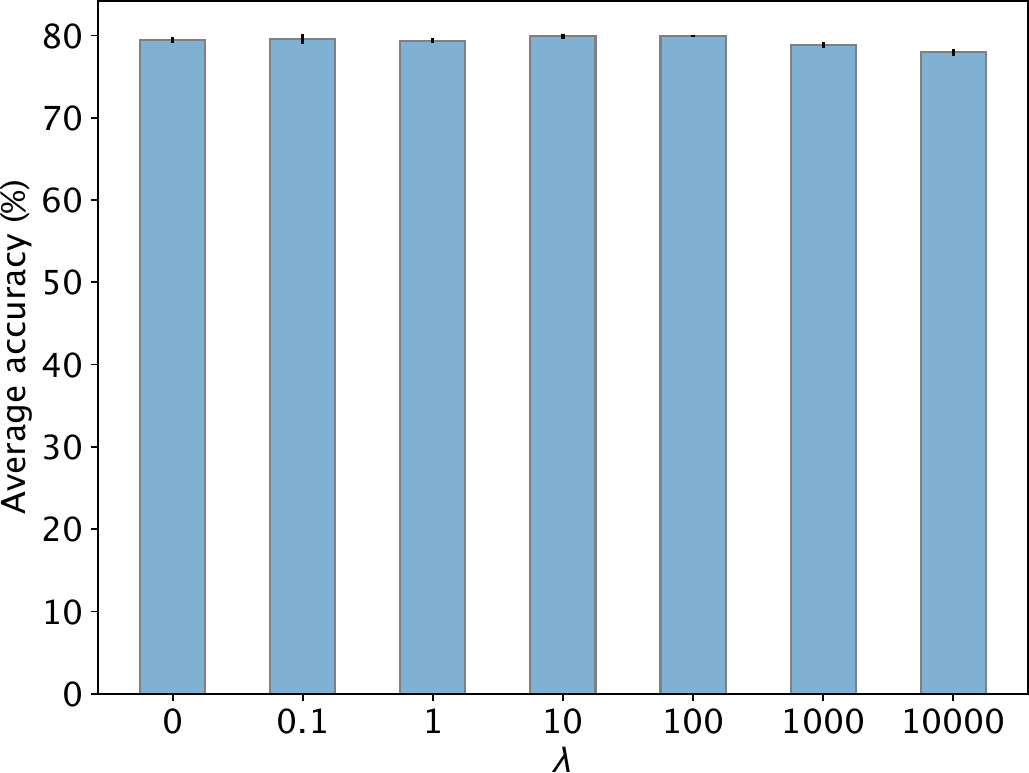}
         \caption{Digits}
         \label{fig:digits-lambda}
     \end{subfigure}
     \hfill
     \begin{subfigure}[b]{0.4\linewidth}
         \centering
         \includegraphics[width=\linewidth]{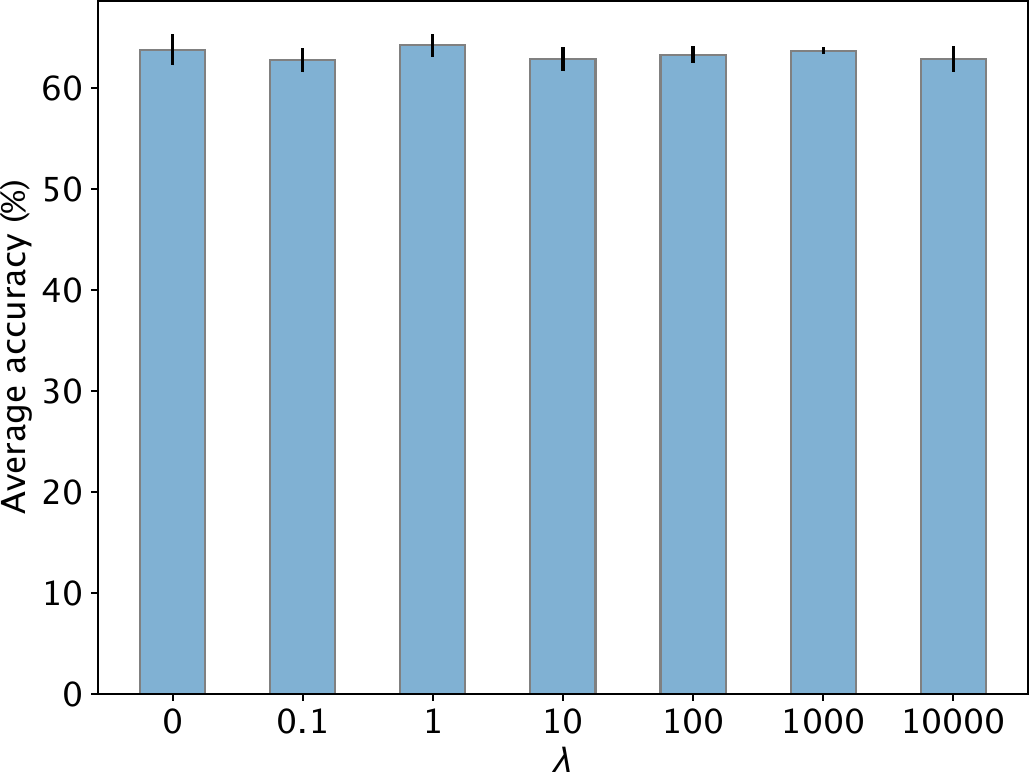}
         \caption{PACS}
         \label{fig:pacs-lambda}
     \end{subfigure}
        \caption{Sensitivity analysis on $\lambda$. We train models with AdvST under different values of $\lambda$. For each $\lambda$, we report average classification accuracy (blue bars) and its standard deviation (vertical black bars) over all target domains for each dataset.}
        \label{fig:sensitivity-analysis-on-lambda}
\end{figure}

The parameter $\lambda$ in Eq. \eqref{eq:objective} controls the size of the uncertainty set. A small $\lambda$ allows the uncertainty set to have distributions with large distributional shifts from the source. With semantics transformations, the average performance of a AdvST-trained model does not change too much under different values of $\lambda$ on the Digits (a maximum drop of 1.94\% in Figure \ref{fig:sensitivity-analysis-on-lambda} (a)) and the PACS (a maximum drop of 1.47\% in Figure \ref{fig:sensitivity-analysis-on-lambda} (b)) datasets. In practice, using a small $\lambda$, e.g., $\lambda=1$ or $\lambda=10$, works well.
\subsubsection{Semantics transformations} 
We analyze how the 12 standard transformations, which are used to construct semantics transformations, affect the generalization performance of a model on target domains. We adopt the leave-one-out strategy to evaluate the contribution of each standard transformation. Specifically, we remove only one standard transformation at a time and train the model using the semantics transformations constructed with the remaining 11 standard transformations. Then, we calculate the difference in classification accuracy on target domains between the model and the model trained using all 12 standard transformations. 

\begin{figure}
    \centering
    \includegraphics{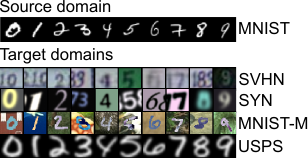}
    \caption{Examples of the Digits dataset.}
    \label{fig:digits-example}
\end{figure}

\begin{figure}[htbp]
    \centering
    \includegraphics[width=0.8\linewidth]{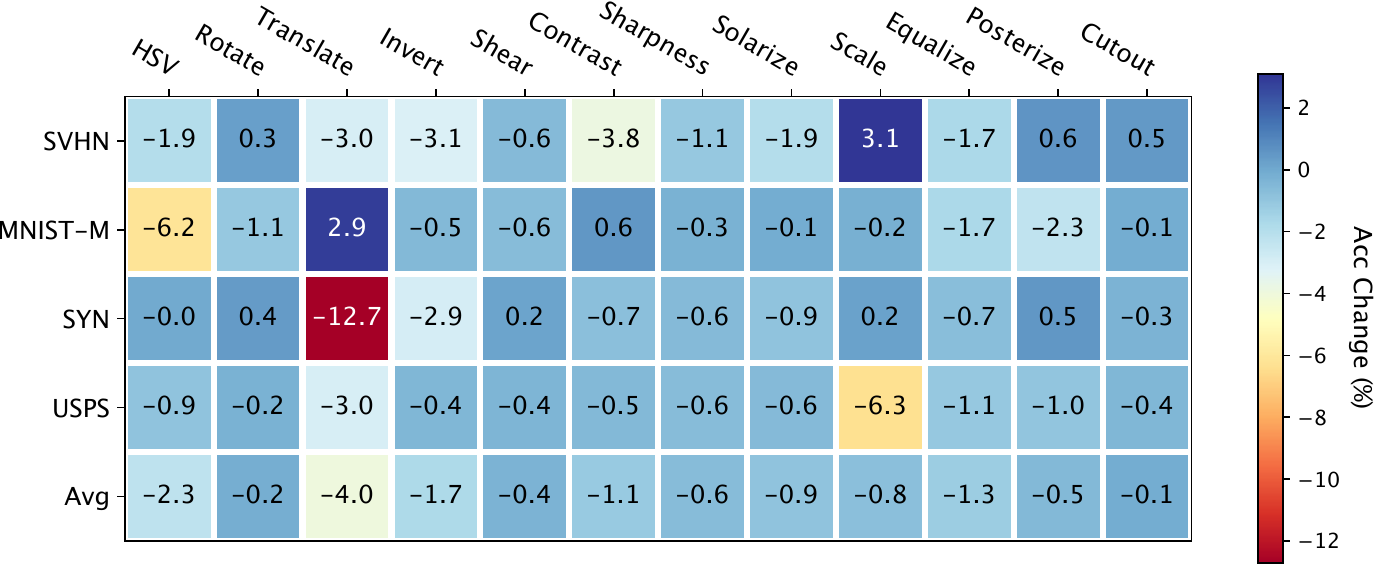}
    \caption{Heatmap of classification accuracy change on the four target domains and average accuracy change after removing one standard transformation (shown as column name).}
    \label{fig:semantics_digits_heatmap}
\end{figure}

We use AdvST-ME and the Digits dataset in this experiment and obtain the heatmap of classification accuracy change in Figure~\ref{fig:semantics_digits_heatmap}. We observe that removing any standard transformation results in a drop in performance. In particular, without \texttt{Translate}, the model has the most drop in average accuracy due to the significant performance drop in the SYN domain, indicating that translational invariance is important for generalizing to the SYN domain. However, removing \texttt{Translate} benefits generalizing to the MNIST-M domain. The contradictory effect of \texttt{Translate} in MNIST-M and SYN domains explains the performance tradeoff between AdvST and AdvST-ME on the two domains observed in Table \ref{tab:digits}. We also observe a similar contradictory effect of  \texttt{Scale} in SVHN and USPS domains and the corresponding performance tradeoff  on the two domains in Table \ref{tab:digits}.
Moreover, we identify that \texttt{Contrast}, \texttt{HSV}, \texttt{Translate}, and \texttt{Scale} are the most important standard transformations for generalizing to the SVHN, MNIST-M, SYN, and USPS domains, respectively.  For example, Scale is beneficial for generalizing to the USPS domain since digits in this domain are enlarged compared to those in the MNIST domain (see examples in Figure~\ref{fig:digits-example}). 

The above analysis highlights the key advantage of our method: semantics transformations can bring domain knowledge, such as translational invariance or scale invariance, that benefits generalization on unseen target domains. Adding more semantics transformations could benefit generalization on target domains; however, it may also bring undesired semantics transformations that have adverse effects on specific target domains.

\subsubsection{In-distribution accuracy} We show the in-distribution accuracy comparison between ERM and our methods on three datasets in Table \ref{tab:in-distribution-cmp} below. Our approach does not hurt the nominal accuracy and slightly improves it.
\begin{table}[h]
\centering
\resizebox{\linewidth}{!}{%
\begin{tabular}{cccc}
\hline
Method & MNIST & Photo (PACS) & Real (DomainNet) \\ \hline
ERM & 98.8 & 98.5 & 76.0 \\
AdvST (ours) & \textbf{99.0} & 99.7 & \textbf{76.7} \\
AdvST-ME  (ours) & 98.9 & \textbf{99.9} & 76.5 \\ \hline
\end{tabular}%
}
\caption{In-distribution accuracy comparison. Models are trained and tested in the same domain.}
\label{tab:in-distribution-cmp}
\end{table}

\begin{figure}[htbp]
    \centering
    \includegraphics[width=0.6\linewidth]{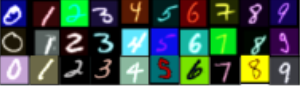}
    \caption{Visualization of the images generated by AdvST for the MNIST domain. }
    \label{fig:mnist-generated}
\end{figure}

\begin{figure}[htbp]
    \centering
    \includegraphics[width=0.6\linewidth]{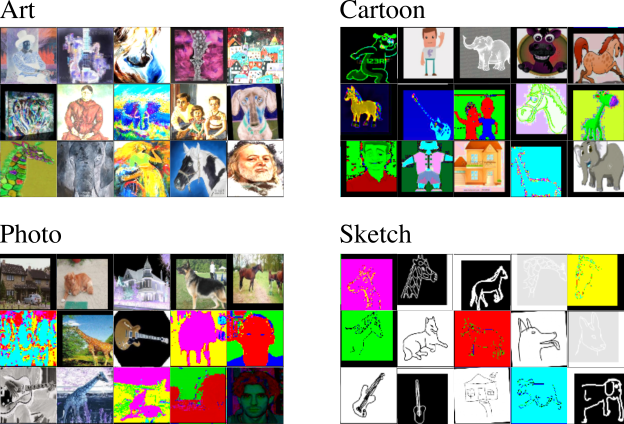}
    \caption{Visualization of the images generated by AdvST for the four domains in the PACS dataset. }
    \label{fig:pacs-visualization}
\end{figure}

\begin{figure}[htbp]
    \centering
    \includegraphics[width=0.6\linewidth]{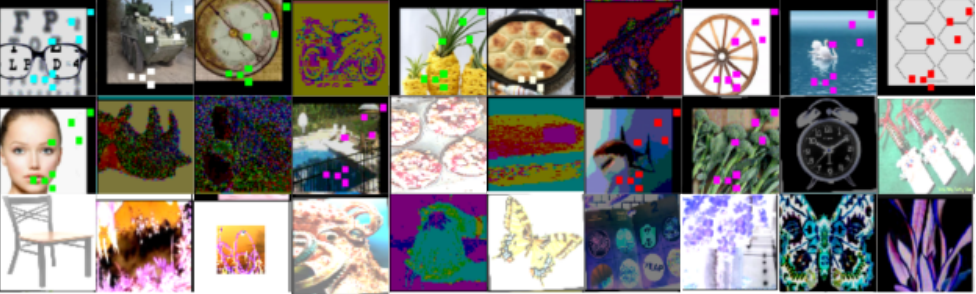}
    \caption{Visualization of the images generated by AdvST for the Real domain in the DomainNet dataset. }
    \label{fig:real-generated}
\end{figure}

\subsubsection{Visualization of generated samples}
We visualize the images generated by AdvST. Figure~\ref{fig:mnist-generated} shows the images generated for the MNIST domain.  Figure~\ref{fig:pacs-visualization} shows the images generated for the four domains in the PACS dataset.  Figure~\ref{fig:real-generated} shows the images generated for the Real domain in the DomainNet dataset. From the three figures, we observe diverse variations in the generated images. For example, images from the Sketch domain in the PACS dataset all have white background and black strokes, while the generated images have various  background and stroke colors.


\subsubsection{Experiments on OfficeHome}
We evaluate our two implementations, AdvST and AdvST-ME on OfficeHome \cite{venkateswara2017deep} which contains four domains (Art, Clipart, Product,and Real) with 65 classes. This is one of the canonical domain adaptation/generalization benchmarks. We use a ResNet-18 \cite{he2016deep} as the backbone network which is pre-trained on ImageNet. The classification layer is a linear layer with 512 inputs and 65 outputs. To calculate the contrastive loss, we design the projection function $\phi$ in Eq.~\eqref{eq:sup-con} as a linear layer with 512 inputs and  128 outputs followed by a normalization layer. We set $\eta=0.01$ and $\epsilon=1$. Other settings are the same as in the PACS experiments. The SDG results are shown in Table \ref{tab:additional-results}.

\begin{table*}[t]
\centering
\caption{Single domain generalization results on OfficeHome. We report the average classification accuracy over the remaining domains when one domain is used as the source domain.}
\label{tab:additional-results}
\begin{tabular}{cc|ccc|cc}
\hline
\multicolumn{2}{c|}{Source} & ADA & ME-ADA & L2D & AdvST & AdvST-ME \\ \hline
\multirow{5}{*}{OfficeHome} & Art & 48.3 & 49.5 & 52.1 & \textbf{52.1}$\boldsymbol{\pm}$\textbf{0.5} & 51.6$\pm$0.3 \\
 & Clipart & 46.1 & 46.9 & 51.2 & \textbf{52.3}$\boldsymbol{\pm}$\textbf{0.3} & 52.0$\pm$0.3 \\
 & Product & 43.9 & 44.3 & 49.4 & \textbf{49.6}$\boldsymbol{\pm}$\textbf{0.3} & 49.2$\pm$0.3 \\
 & Real & 53.6 & 54.6 & 58.2 & \textbf{60.1}$\boldsymbol{\pm}$\textbf{0.2} & 59.9$\pm$0.4 \\
 & Avg. & 48.0 & 48.8 & 52.7 & \textbf{53.5}$\boldsymbol{\pm}$\textbf{0.1} & 53.2$\pm$0.2 \\ \hline
\end{tabular}%
\end{table*}

\end{document}